\documentclass{article}

\usepackage{microtype}
\usepackage{graphicx}
\usepackage{subcaption}
\usepackage{booktabs} 

\usepackage{hyperref}

\usepackage[accepted]{icml2018}
\usepackage{tabularx}
\usepackage{todonotes}
\usepackage{amsmath}
\usepackage{amssymb}
\usepackage{mathtools}
\usepackage{latexsym}
\usepackage{algorithm,algorithmic}

\DeclareMathOperator*{\argmax}{arg\,max}

\icmltitlerunning{Automatic Configuration of Deep Neural Networks with EGO}

\begin{document}

\twocolumn[
\icmltitle{Automatic Configuration of Deep Neural Networks\\
	 with Parallel Efficient Global Optimization}



\icmlsetsymbol{equal}{*}

\begin{icmlauthorlist}
\icmlauthor{Bas van Stein}{equal,to}
\icmlauthor{Hao Wang}{equal,to}
\icmlauthor{Thomas B\"ack}{to}
\end{icmlauthorlist}

\icmlaffiliation{to}{LIACS, University of Leiden, Leiden, The Netherlands}
\icmlcorrespondingauthor{Bas van Stein}{b.van.stein@liacs.leidenuniv.nl}
\icmlcorrespondingauthor{Hao Wang}{h.wang@liacs.leidenuniv.nl}

\icmlkeywords{Bayesian Optimization, Deep Learning, Network Architectures}

\vskip 0.3in
]



\printAffiliationsAndNotice{\icmlEqualContribution} 

\begin{abstract}
Designing the architecture for an artificial neural network is a cumbersome task because of the numerous parameters to configure, including activation functions, layer types, and hyper-parameters. With the large number of parameters for most networks nowadays, it is intractable to find a good configuration for a given task by hand.
In this paper an Efficient Global Optimization (EGO) algorithm is adapted to automatically optimize and configure convolutional neural network architectures. A configurable neural network architecture based solely on convolutional layers is proposed for the optimization. Without using any knowledge on the target problem and not using any data augmentation techniques, it is shown that on several image classification tasks this approach is able to find competitive network architectures in terms of prediction accuracy, compared to the best hand-crafted ones in literature. In addition, a very small training budget (200 evaluations and 10 epochs in training) is spent on each optimized architectures in contrast to the usual long training time of hand-crafted networks. Moreover, instead of the standard sequential evaluation in EGO, several candidate architectures are proposed and evaluated in parallel, which saves the execution overheads significantly and leads to an efficient automation for deep neural network design.
\end{abstract}

\section{Introduction}
\label{sec:Intro}

Deep Artificial Neural Networks and in particular Convolutional Neural Networks (CNN) have demonstrated great performance on a wide range of difficult computer vision, classification and regression tasks. One of the most promising aspects of using deep neural networks is that feature extraction and feature engineering, which was mostly done by hand so far, now is completely taken care of by the networks themselves.
Unfortunately, the design and configuration of the artificial neural networks are still derived by hand using either an educated guess, by popularity (using an architecture from previous literature) or by trying a grid of different architectures and parameters and then choosing the best performing network.
Since the number of choices for a network architecture and its parameters can become quite large, an optimal deep neural network for a given problem is very unlikely to be obtained using this hand-crafted procedure.

The challenges in configuring CNNs are: 1) the search space is usually high dimensional and heterogeneous, resulting from a large number of structure choices (e.g., number of layers, layer type, etc.) and real parameters. 2) the computational time becomes the bottleneck when fitting a deep network structure to a relatively large data set. Those difficulties hinder the applicability of the traditional nonlinear black-box optimizers, for instance Evolutionary Algorithms~\cite{stanley2002evolving}. Instead, it is proposed here to adopt the so-called \emph{Efficient Global Optimization}~\cite{movckus1975bayesian,mockus2012bayesian,jones1998efficient} (EGO) algorithm as the network configurator. The standard EGO algorithm is a sequential strategy designed for the expensive evaluation scenario, where a single candidate configuration is provided in each iteration. It is proposed to adapt the EGO algorithm to yield several candidate configurations in each iteration where the resulting configurations can be evaluated in parallel. 

This paper is organized as follows. In section 2, the related approaches on network configuration are discussed. In section 3, we introduce the All-CNN configuration framework, using only convolutional layers, and the EGO-based configurator is explained in section 4. The proposed method is validated and tested in sections 5 and 6, followed by the demonstration of an application on a real-world problem.


\section{Related Research}
\label{sec:Related}
The optimization of hyper-parameters is a very known challenge and has been addressed in many works.
For example, \cite{bergstra2012random} shows that random chosen trials are more efficient than using grid search to perform hyper-parameter optimization.
Obviously, both random and grid search are far from optimal, and more sophisticated methods are required to search the very large and complex search space for optimizing deep artificial neural networks.
More recent work of the same author \cite{bergstra2013making} shows that automatic hyper-parameter tuning can yield state-of-the-art result, In these papers, architectures are used that are known to work on a specific problem and are then fine-tuned by hyper-parameter optimization.
Some other sophisticated algorithms to perform parameter tuning and automated machine learning configuration are Bayesian Optimization \cite{snoek2012practical,jones1998efficient}, Evolutionary Algorithms \cite{loshchilov2016cma} and SMAC \cite{Hutter2011}, which try to quickly converge to practical well-performing hyper-parameters for a given machine learning algorithm.

Unfortunately, even with these sophisticated algorithms, optimization of the deep neural network architecture itself, in addition with its hyper-parameters, is a very challenging task. This is caused by the time complexity and computational effort that is required to train these networks, in combination with the size of the search space of hyper-parameters for such networks.
Automatically optimizing the structure of an artificial neural network is not an entirely new idea though, as already in 1989 \cite{miller1989designing} genetic algorithms were proposed to optimize the links between a predefined number of nodes.
A bit later, an evolutionary program (GNARL) was proposed to evolve the structure of recurrent neural networks \cite{angeline1994evolutionary}. 
In another, more recent work \cite{ritchie2003optimizationof}, \emph{Genetic Programming} (GP) is used for the automatic construction of neural network architectures.

One of the main bottlenecks with the already proposed methods though is that a single evaluation of an artificial neural network can take several hours, on a modern GPU system. This makes it infeasible to apply these algorithms with a large evaluation budget or on a large problem instance. Unfortunately, these algorithms usually require a large evaluation budget to find well performing network configurations for a specific problem.
Another challenge is to define a bounded search space that still covers most of the possibilities in order to find the optimum. When dealing with neural network structures this is far from simple. The number of layers for example could be a problematic parameter to vary, since each layer comes with its own set of hyper-parameters. 

To alleviate this problem, a generic configurable deep neural network architecture is proposed in this paper. This architecture is highly configurable with a large number of parameters and can represent very shallow to very deep convolutional neural networks. The configurable architecture has a fixed number of hyper-parameters and is therefore very suitable for optimization.
To tackle this optimization task, the well-known \emph{Efficient Global Optimization} algorithm~\cite{movckus1975bayesian,mockus2012bayesian,jones1998efficient} is adopted with several important improvements, enabling the parallel training of different network candidates. The main advantages of the proposed approach are:
\begin{enumerate}
\item Small optimization time: it requires by far less real evaluations (training of candidate networks) than other approaches.
\item Parallelism: several candidate networks are suggested in each iteration, facilitating parallel execution over multiple GPUs.
\end{enumerate}

\medskip

\section{A Configurable All-Convolutional Neural Network}
\label{sec:network}
In order to optimize the structure and hyper-parameters of a deep neural network, a few modeling decisions are required to set the boundaries of the search space. 
The complexity of the search space is mostly due to a large number of different layer types, activation functions and regularization methods, each coming with their own set of hyper-parameters.

In order to reduce the complexity of the search space without making too many modeling assumptions, a generic configurable convolutional neural network designed for any image classification problem is proposed here.

According to \cite{springenberg2014striving}, using only convolutional neural network layers can give the same or better performance as using the often used structure of convolutional layers followed by a pooling layer. Therefore, for our generic configurable network structure, we have chosen to use only convolution layers with the exception of the final layer. 

\begin{figure}[!ht]
    \centering
    	\includegraphics[width=\linewidth, trim=50mm 20mm 50mm 0mm, clip]{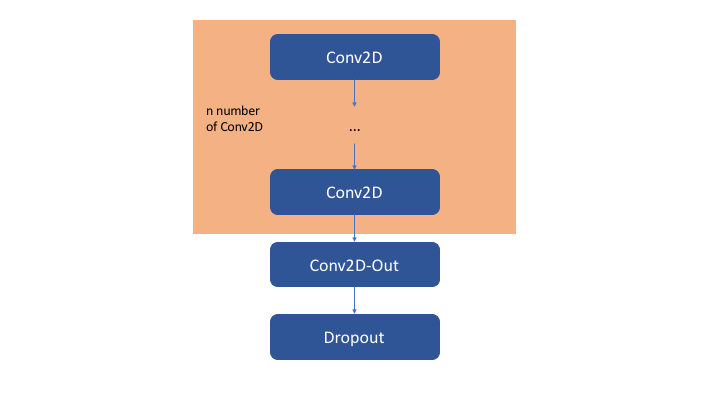}
    	\caption{Schematic diagram of the stack structure.\label{stack}}
\end{figure}

The configurable network architecture is shown in Table \ref{tab:generic} where each of the $q$ stacks has an architecture as shown in Figure \ref{stack}. The network consists of multiple of these stacks, that each consist of a number of convolutional layers and a convolutional layer with strides (Conv2D-Out), to allow for pooling, and a dropout layer. The last part of the network uses either global average pooling or not, and ends in a dense layer with the size of the number of classes one wants to predict. 
Each stack has $7$ independent configurable parameters and $2$ shared parameters that can be optimized.
The convolutional layers in the stack have the parameters $f, k, l2$ and $s$, which are the number of filters $f$, the kernel size $k$, the $l2$ regularization factor for the weights, and for the Conv2D-Out layer the strides $s$, respectively.  The parameter $a$ stands for the configurable activation function and every dropout has its own dropout probability ($d$).  The last dense layer has $l2$ and $a$ as configurable parameters. The size of each stack is configurable as well ($n$), and allows for very shallow to very deep neural network architectures. All hyper-parameters that are not taken into account for the configuration are set to the values recommended by literature and the padding for each convolution layer is set to `same' in order to avoid negative dimensions. 

\begin{table}[!h]
\caption{Generic Configurable All-CNN structure with $q$ stacks and configurable parameters per layer. $i$ is the index for the current stack ($i = 1, \dots , q$). \label{tab:generic}}
\centering
\begin{tabularx}{\linewidth}{X|r}
\toprule
\textbf{Layer Type} & \textbf{Parameters} \\
\midrule

Dropout & $d_0$ \\
Conv2D  & $f_0, k_0, l2, a$\\
\midrule
	\multicolumn{2}{@{}l@{}}{$q$ Stacks} \\[0.25ex]
\midrule
$n_i \times $ Conv2D  & $f_i, k_i, l2, a, n_i$\\
Conv2D-Out  & $f_{out-i}, k_{out-i}, s_{out-i}, l2, a$\\
Dropout & $d_i$ \\
\midrule
	\multicolumn{2}{@{}l@{}}{Head} \\[0.25ex]
\midrule
GlobalPooling 		& boolean \\
Dense 			 	& $l2, a_{out}$ \\
\bottomrule
\end{tabularx}
\end{table}

Next to the parameters of the configurable network itself (which are $26$ when using $3$ stacks), there are the learning rate ($lr$) and decay rate $(\lambda)$ for the back-propagation optimizer.
Depending on available resources and the classification task at hand, the ranges of the parameters can be determined by the user.
For this paper, the ranges can be found in Table \ref{tab:ranges}. The optimizer used for back-propagation is the well-known stochastic gradient descent (SGD), provided by the Keras \cite{chollet2015keras} python library.

\begin{table}[!h]
\caption{Ranges of the search space dimensions} \label{tab:ranges}
\centering
\begin{tabular}{@{}lr@{}}
\toprule
Parameter & Range \\
\midrule
$a$ & \{elu, relu, tanh, selu, sigmoid\} \\
$f$ & $[1 .. 512]$ \\
$k$ & $[1 .. 8]$ \\
$s$ & $[1 .. 5]$ \\
$n$ & $[1 .. 6]$ \\
$d$ & $[10^{-5}, 0.8]$ \\
$l2$ & $[10^{-5}, 10^{-2}]$ \\
$lr$ & $[10^{-5}, 1.0]$ \\
\bottomrule
\end{tabular}

\end{table}

\section{Efficient Global Optimization based Configurator}
\label{sec:bayes}
The search space of the All-CNN framework is heterogenous and high dimensional. For the integer parameters, in case of three stacks, there are seven for the number of filters ($f$), seven for the kernel size ($k$), three for strides ($s$) and three for the number of layers ($n$) in the stack, and thus $20$ in total. For the discrete parameters, there are two for the activation functions ($a$) of the stack and the head and for the real parameters, there are four parameters for the dropout rate ($d$), one for regularization ($l2$) and one for the learning rate ($lr$). In addition, we have one boolean variable to control the global pooling. Therefore, this search space can be represented as:
$$\mathcal{C} = \mathbb{R}^{6} \times \mathbb{Z}^{20} \times \mathcal{B} \times \mathcal{D}^2,$$
where $\mathcal{B}=\{0, 1\}$ and $\mathcal{D}=\{\text{elu}, \text{relu}, \text{tanh}, \text{selu}, \text{sigmoid}\}$. The convolutional neural network can be instantiated by drawing samples in $\mathcal{C}$. Given a data set, the problem arises in finding the optimal configuration, 
with respect to a pre-defined, real-valued performance metric $f$ of the neural network (for instance, $f$ can be set to $r^2$ for regression tasks and precision for classification problems): 
$f : \mathcal{C} \rightarrow \mathbb{R}.$
In the following discussion, it is assumed that the performance metric $f$ is subject to minimization, without loss of generality (the maximization problem can be easily converted).
The challenge in optimizing $f$ is the evaluation time of itself, which will be extremely expensive when training a large network structure on a huge data set. Consequently, it is recommended to use efficient optimization algorithms that can save as many evaluations as possible. The efficient global optimization (EGO) algorithm~\cite{movckus1975bayesian,mockus2012bayesian,jones1998efficient} is a suitable candidate algorithm for this task. It is a \emph{sequential} optimization strategy that does not require the derivatives of the objective function and is designed to tackle expensive global optimization problems. Compared to alternative optimization algorithms (or other design of experiment methods), the distinctive feature of this method is the usage of a meta-model, which gives the \emph{predictive distribution} over the (partially) unknown function. 

Briefly, this optimization method iteratively proposes new candidate configurations over the meta-model, taking both the prediction and model uncertainty into account. After the evaluation of the new candidate configurations, the meta-model will be re-trained.

\subsection{Initial Design and Meta-modeling}
To construct the meta-model, some initial samples in the configuration space, $X=\{{\mathbf{x}^{(1)}},\mathbf{x}^{(2)}, \ldots, \mathbf{x}^{(n)}\} \subset \mathcal{C}$ are generated via the Latin hypercube sampling (LHS) ~\cite{10.2307/1268522}. The corresponding performance metric values are obtained by instantiating the network and validating its performance on the data set: $Y=\{y^{(1)},y^{(2)},\ldots,y^{(n)}\}=\{f(\mathbf{x}^{(1)}),f(\mathbf{x}^{(2)}),\ldots,f(\mathbf{x}^{(n)})\}$. Note that the evaluation of the initial designs can be easily parallelized. For the choice of meta-models, although Gaussian process regression~\cite{Sacks1989,santner2003design} (referred to as Kriging in geostatistics~\cite{Krige51}) is frequently used in EGO, we adopt the \emph{random forest} instead, due to the fact that it is more suitable for a mixed integer configuration domain~\cite{hutter2011sequential}. In the following discussions, the prediction on configuration $\mathbf{x}$ is denoted as $m(\mathbf{x})$. In addition, the empirical variance of the prediction $\hat{s}^2(\mathbf{x})$ is also calculated from the forest, which quantifies the prediction uncertainty.

\subsection{Infill-Criterion} 
To propose potentially good configurations in each iteration, the so-called infill-criterion is used to quantify the quality criterion of the configurations. Informally, infill-criteria work in a way that predicted values from the meta-model and the prediction uncertainty are balanced. A lot of research effort has been put over the last decades in exploring various infill-criteria, e.g., Expected Improvement~\cite{movckus1975bayesian,jones1998efficient}, Probability of Improvement~\citep{jones2001taxonomy,vzilinskas1992review} and Upper Confident Bound~\cite{auer2002using,Srinivas2010}. In this contribution, we adopt the so-called Moment-Generating Function (MGF) based infill-criterion, as proposed in~\cite{wang.smc17}. This infill-criterion allows for explicitly balancing exploitation and exploration. This criterion has a closed form and can be expressed as:
\begin{align}
	\mathcal{M}(\mathbf{x};t)
	&=\Phi\left(\frac{y_{\text{min}} - m'(\mathbf{x})}{s(\mathbf{x})}\right) \nonumber \\ \cdot &\exp\left(\left(y_{\text{min}} - m(\mathbf{x}) - 1\right)t + \frac{\hat{s}^2(\mathbf{x})}{2}t^2\right) 
	\label{eq:MBA} \\
	m'(\mathbf{x}) &= m(\mathbf{x}) - \hat{s}^2(\mathbf{x})t, \nonumber
\end{align}
where $y_{\text{min}} = \min\{y^{(1)},y^{(2)},\ldots,y^{(n)}\}$ is the current best performance over all the evaluated configurations and $\Phi(\cdot)$ stands for the cumulative distribution function of the standard normal distribution. The infill-criterion $\mathcal{M}$ introduces an additional real parameter $t$ (``temperature'') to explicitly control the balance between exploration and exploitation. As explained in~\cite{wang.smc17}, when $t$ goes up, $\mathcal{M}$ tends to reward the configurations with high uncertainty. On the contrary, when $t$ is decreased, $\mathcal{M}$ puts more weight on the predicted performance value. It is then possible to set the $t$ value according to the budget on the configuration task: with a larger budget of function evaluations, $t$ can be set to a relatively high value, leading to a slow but global search process and vice versa for a smaller budget.

\subsection{Parallel execution}
Due to the typically large execution time of instantiated network structures, it is also proposed here to parallelize the execution. This requires generating more than one candidate configuration in each iteration. Many methods are developed for this purpose, including multi-point Expected Improvement~\cite{Ginsbourger2010} and Niching techniques~\cite{wang.evolve15}. Here, we adopt the approach in~\cite{hutter2012parallel}, where $q$ ($>1$) different temperatures $t$ are sampled from the log-normal distribution $Lognormal(0, 1)$ and $q$ different $\mathcal{M}$ criteria are instantiated using the temperatures accordingly. Consequently, $q$ candidate configurations can be obtained by maximizing those $q$ infill-criteria. On one hand, as log-normal is a long-tailed distribution, most of the $t$ values are realized relatively small and thus the model prediction is well exploited. On the other hand, only a few $t$ samples will be relatively high and therefore will lead to very explorative search behavior. 

To maximize the infill-criterion on the mixed-integer search domain, we adopt the so-called Mixed-Integer Evolution Strategy (MIES)~\cite{li2013mixed}. The proposed Bayesian configurator is summarized in Algorithm~\ref{AL:ego}.
\begin{algorithm} 
	\caption{EGO Configurator}
	\label{AL:ego}
	\begin{algorithmic}[1]
		\STATE{Generate the initial design $X$ using LHS.}
		\STATE{Construct the initial random forest on $(X,Y)$.}
		\WHILE{the stopping criterion is not fulfilled}
			\FOR{$i = 1 \rightarrow q$}
				\STATE{$t_i \leftarrow Lognormal(0, 1)$}
				\STATE{Maximize the infill-criterion using Mixed integer Evolution Strategy: $$\mathbf{x}_i' = \argmax_{\mathbf{x}\in \mathcal{C}}\mathcal{M}(\mathbf{x}; t_i)$$}
			\ENDFOR
			\STATE{Parallel training and performance assessment for all $\{\mathbf{x}_i'\}_{i=1}^{q}$: $y_i' \leftarrow f(\mathbf{x}_i')$.}
	 		\STATE{Append $\{\mathbf{x}_i', y_i'\}_{i=1}^{q}$ to $(X, Y)$.}
	 		\STATE{Re-train the random forest model of $f$ on the augmented data set $(X, Y)$}
	 	\ENDWHILE
	\end{algorithmic}
\end{algorithm}

\section{Experiments}
\label{sec:Experiments}

To test our algorithm, two very popular and common classification tasks have been performed using the proposed configurator and a configurable network with $3$ stacks.
These are the \emph{MNIST} dataset \cite{lecun1998gradient}, containing 60.000 training samples, and a test set of 10.000 examples, all 28x28 greyscale images, and the \emph{CIFAR-10} dataset \cite{krizhevsky2009learning}, containing 60.000, 32x32 colour images with 10 classes, divided into 6000 images per class. There are 50.000 training images and 10.000 test images, in this case.

In the optimization procedure of the neural network on the MNIST dataset, each evaluation is run for $10$ epochs only with a batch size of $100$ images. For the CIFAR-10 dataset, the number of epochs is increased to $50$, which is still much less than the number of epochs in most recent literature ($> 300$).
An early stopping criterion is used to stop the evaluation of a particular configuration after $6$ epochs of no improvements. No data augmentation is used.

The Bayesian mixed integer configurator is set to evaluate $5$ network configurations per step in parallel using NVIDIA K80 GPUs, where the first $5$ steps are used for the initial LHS design. The test set accuracy is returned after each evaluation as fitness value for the optimizer. 


\section{MNIST and CIFAR-10 Results}

\begin{figure*}[!ht]
    \centering
    \begin{subfigure}[t]{0.49\textwidth}
    	\includegraphics[width=\linewidth]{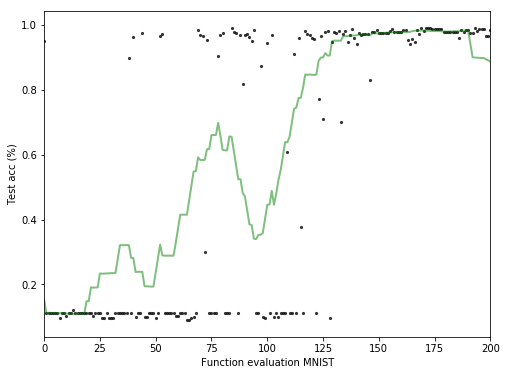}
    	\caption{MNIST, $200$ evaluations}
    \end{subfigure}
    \begin{subfigure}[t]{0.49\textwidth}
    	\includegraphics[width=\linewidth]{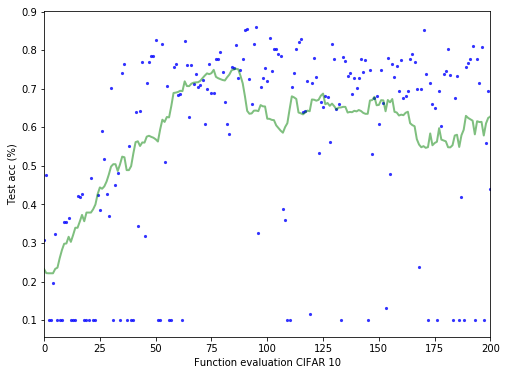}
    	\caption{CIFAR-10, $200$ evaluations}
    \end{subfigure}
    \caption{a) The left plot shows the optimization run on MNIST, plotting the test accuracy (black dots) of $200$ evaluations using $10$ epochs. The $20$-moving average is depicted by the green line.  b) The right plot shows similar results on the CIFAR-10 classification task using $50$ epochs per evaluation. \label{fig:mnistcifar}}
\end{figure*}

In Figure \ref{fig:mnistcifar} the results of the automatic configuration of the All-CNN networks are shown. In both cases, after approximately $50$ evaluations, a well-performing network configuration is obtained. Both classification tasks used exactly the same initial configuration, the only difference is the number of epochs for each network evaluation.

The best performing configurations compete with the state-of-the-art as shown in Table \ref{tab:mnist} and Table \ref{tab:cifar} and can be possibly improved when trained using more epochs. It should be noted that the number of epochs used to obtain these results is significantly lower than the number of epochs in state-of-the-art solutions from literature. The advantage of such a small number of epochs is that it speeds up the entire optimization process. The idea behind this is that well-performing configurations can be tuned with a larger number of epochs as second optimization step, most likely resulting in increased performance.
Using the automatic configurator we obtained neural network architectures that compete with state-of-the-art results using only $200 \cdot 10$ epochs and $200 \cdot 50$ epochs in total, without any manual tuning, reconfiguring or upfront knowledge of the specific problem instances. While hand-crafted network configurations are not only trained using many more epochs for the final reported architecture, they also require a huge amount of time to be constructed by reconfiguring and fine-tuning the architecture. Therefore, the hand-crafted networks basically use many more more epochs until the final architecture is reached.

\begin{table}[!h]
\caption{MNIST Performance from literature.} \label{tab:mnist}
\centering
\begin{tabular}{@{}llr@{}}
\toprule
Test error & Algorithm  & Epochs \\
\midrule
$0.23\%$ & \cite{ciregan2012multi} & $800$  \\
$0.32\%$ &  \cite{graham2014fractional} & $250$  \\
$0.61\%$ & \textbf{Optimized All-CNN} & $10$ \\
$0.71\%$ & \cite{yang2015deep} & unknown \\
\bottomrule
\end{tabular}

\end{table}

\begin{table}[!h]
\caption{CIFAR-10 Performance from literature.} \label{tab:cifar}
\centering
\begin{tabular}{@{}llr@{}}
\toprule
Accuracy & Algorithm & Epochs \\
\midrule
$95.59\%$ &	\cite{graham2014fractional}  & $250$  \\
$95.59\%$ &	\cite{springenberg2014striving} & $350$  \\
$86.46\%$ & \textbf{Optimized All-CNN} & $50$  \\
$84.87\%$  & \cite{zeiler2013stochastic} & $500$	\\
\bottomrule
\end{tabular}

\end{table}

\section{Real World Problem: Tata Steel}

The proposed algorithm is applied on the real world problem of classifying defects during the hot rolling process of steel.
This industrial process is very complex with many conditions and parameters that influence the final product. It is also a process that changes over the years and requires dealing with concept shift and concept drift. 
One of the main objectives for Tata Steel is to automatically classify and predict surface defects using material properties and machine parameters as input data. 
To achieve this objective, first, a deep neural network architecture is designed by hand to classify these defects.

The Tata Steel data set consists of various material measurements and machine parameters. Most of the measurements are measured over the complete length of each coil but not over the width of the coil (since the width is much smaller). However, the temperature measurements are taken over several tracks in the width of the coil as well. Due to this spatial difference, it was decided to design two concatenated network architectures. One part of the architecture is based purely on the temperature data, allowing for the application of convolution layers in the width and length direction of the coil. The second component is used for modeling the remaining measurements and machine parameters where the convolution filters only work in the length of the coil. In the end of the design process, these two parts are merged into one final fully-connected output layer. 

The initial design process of these architectures was mainly based on trial and error and recommendations from literature. The design process started with a small, relative simple, two-layer multi-perceptron, and adding additional dense and convolution layers in order to increase the final accuracy. Dropout is being applied to prevent over-fitting, and after several manual iterations a dropout rate of $0.2$ seemed to work best. 

Next, we applied a slightly modified version of the proposed configurable all-CNN network (with a separate stack for the temperature data before concatenating it to the main model) and automatically optimized the configuration. The optimal configuration obtained by using our optimization procedure significantly improves the classification accuracy. It also allows for easy retraining and validation on future data, since almost zero knowledge of the actual dataset is required to train and optimize the network architecture.

\begin{figure}[!ht]
    \centering
    \begin{subfigure}[t]{0.48\linewidth}
    	\includegraphics[width=\linewidth]{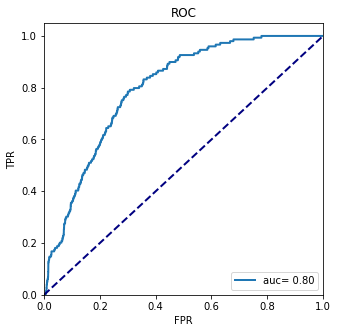}
    	\caption{ROC of hand-constructed classifier.}
    \end{subfigure}
    \begin{subfigure}[t]{0.48\linewidth}
    	\includegraphics[width=\linewidth]{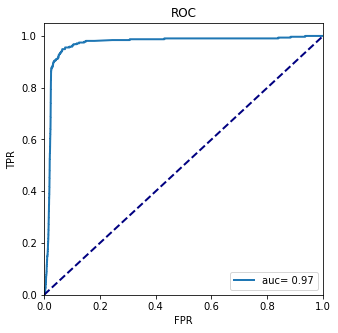}
    	\caption{ROC of optimized All-CNN.}
    \end{subfigure}
    \caption{a) The left plot shows the ROC curve of the classification task for Tata Steel with a hand-crafted neural network architecture.  b) The right plot shows the ROC curve for the same task but now using the optimized neural network architecture using $200$ evaluations for the optimization and only $5$ epochs per evaluation. \label{fig:tata}}
\end{figure}

The test set accuracy of the hand-designed classifier and the optimized classifier for this real world application is shown in Figure \ref{fig:tata}. It can be observed that the optimized classifier has a significantly improved accuracy on this specific defect type, with an almost $90\%$ true positive rate with only a very small ($0.5\%$) false positive rate. This shows that the optimization procedure and configurable network architecture has great potential for industrial applications.


\section{Conclusions and Outlook}
\label{sec:conclusions}
A novel approach based on Efficient global optimization algorithm is proposed to automatically configure the neural networks architecture. On some well-known image classification tasks,  it is observed that the proposed optimization approach is capable of generating well-performing networks with a limited number of optimization iterations. In addition, the resulting optimized neural networks are also highly competitive with the state-of-the-art manually designed ones on the MNIST and CIFAR-10 classification task. Note that such performance of the optimized network are achieved under a very small number of epochs ($10$ for MNIST, and $50$ for CIFAR-10) for training, without any knowledge on the classification task or data augmentation techniques.

As for the real-world problem, we have applied the proposed approach on the challenge of steel surface detection. The outcome clearly illustrates that the proposed configuration approach also works extremely well. The accuracy of the optimized network that detects the surface defect for Tata Steel is significantly higher than the accuracy of the network designed by hand, which is obtained with manual fine-tuning.

For the next step, there are several possibilities to work on. First, the proposed approach will be applied and tested on various modeling tasks and real-world problems. Second, the actual training time of the candidate network will be taken into account explicitly. The trade-off between training time and accuracy can be controlled by optimizing the epochs and batch size. Additionally, it is also interesting to formulate this as a bi-criteria decision making problem, with one objective being the accuracy of the network and the other objective the training time required. Third, we will investigate how to extend the current configurable network that has a linear topology, to more general topological structures. In this case, it will be very challenging to search efficiently in the complex configuration space with multiple dependencies.

\paragraph*{Acknowledgment}
The authors acknowledge support by NWO (Netherlands Organization for Scientific Research) PROMIMOOC project (project number: 650.002.001).

\bibliography{main}
\bibliographystyle{icml2018}

\end{document}